%% file: main.tex
\documentclass[runningheads]{llncs}
\usepackage{graphicx}
\usepackage{todonotes}
\usepackage{multirow}
\usepackage{array}
\usepackage{adjustbox}
\usepackage{amstext}
\usepackage{cleveref}
\usepackage[bottom]{footmisc}
\begin{document}
\title{Traffic Flow Forecast of Road Networks with Recurrent Neural Networks}
%
%
\author{
Ralf R\"uther
\and
Andreas Klos
\and
Marius Rosenbaum
\and
Wolfram Schiffmann
}
\authorrunning{R. R\"uther et al.}
%
\institute{FernUniversität in Hagen, Germany\\
\email{ralf.ruether@gmail.com}\\
\email{andreas.klos,marius.rosenbaum,wolfram.schiffmann@fernuni-hagen.de}\\
\url{https://fernuni-hagen.de/rechnerarchitektur/en/}}
\maketitle

\input{abstract.tex}

\input{section1/section1.tex} 
\input{section2/section2.tex}
\input{section3/section3.tex}
\input{section4/section4.tex}

\input{section5/section5.tex}

\bibliographystyle{splncs04}
\bibliography{main.bbl}

\end{document}

%% file: abstract.tex
\begin{abstract}

The interest in developing smart cities has increased dramatically in recent years. In this context an intelligent transportation system depicts a major topic. The forecast of traffic flow is indispensable for an efficient intelligent transportation system. The traffic flow forecast is a difficult task, due to its stochastic and non linear nature. Besides classical statistical methods, neural networks are a promising possibility to predict future traffic flow. In our work, this prediction is performed with various recurrent neural networks. These are trained on measurements of induction loops, which are placed in intersections of the city. We utilized data from beginning of January to the end of July in 2018. Each model incorporates sequences of the measured traffic flow from all sensors and predicts the future traffic flow for each sensor simultaneously. A variety of model architectures, forecast horizons and input data were investigated. Most often the vector output model with gated recurrent units achieved the smallest error on the test set over all considered prediction scenarios. Due to the small amount of data, generalization of the trained models is limited.

\keywords{Deeplearning \and RNN  \and LSTM \and GRU \and Traffic Forecast}
\end{abstract}

%% file: section1/section1.tex
\section{Introduction}

An intelligent transportation system (ITS) occupies a fundamental role regarding the development of smart cities \cite{Wei2019}. Hence, traffic forecast has become an important task for ITSs which facilitates the planning and development of traffic management and control systems \cite{Jingyuan2017}. The necessity of well conceived management and control systems dramatically increases due to the steadily rising number of cars which often causes the traffic infrastructure of cities to collapse especially during rush hours. Furthermore, commuters waste a lot of time by traffic jams and the emissions of their cars pollute the environment \cite{Wei2019}. For that reason, many algorithms have been developed and applied for traffic forecasting. 

Those algorithms have been subdivided in \cite{Lippi2013} into classical time-series and machine learning (ML) approaches. A selection of classical time-series and ML techniques -- comprising artificial neural networks (ANNs) -- were applied to the California Freeway Performance Measurement System (PeMS) dataset. The results are compared to each other with the conclusion that the classical time-series approaches seasonal autoregressive integrated moving average (SARIMA) with a kalman filter, outperforms all other techniques regarding the measured error. Nevertheless, the training time was about 8 $\times$ and the prediction duration more than 12 $\times$ longer compared to the ANN. Unfortunately, only a shallow ANN was analyzed. Furthermore, the forecast horizon was constrained to 15 min.

In \cite{Peng2018} SARIMA is compared to an ANN with two hidden layers. The prediction horizon varies from 1\,h to 24\,h, the input data is restricted to working days (7:00 am - 7:00 pm) and the weather condition is incorperated. The ANN outperformed the SARIMA model regarding the error value. More complex ANN models as well as training and testing time are left unconsidered.

In \cite{Lv2015} stacked autoencoders (SAE) are applied to the PeMS dataset -- only working days -- with various prediction horizons ranging from 15 to 60 min. A grid search was performed, to find well performing SAE architectures. The SAE models had the lowest error values compared to a variety of ML approaches. A comparison to other deeplearning (DL) models is lacking.

Besides, in \cite{Jingyuan2017} a bi-directional long short-term memory (DBL) model is introduced and compared to other  models, like e.\,g. long short-term memory (LSTM), SAE etc. The efficiency is proved with the PeMS dataset. The DBL was able to outperform all considered models regarding the error values. The presented model complexity, the training and testing duration are omitted.

In \cite{Tian2015} an one hidden layer LSTM model is proposed. The same prediction horizons like in \cite{Lv2015} were chosen. The experiments were performed on the PeMS dataset of 2014 for 30 selected observation stations. The analyses showed, that the LSTM achieved the smallest error rate compared to other ML models. Though, the training and testing time are neglected.

Another approach based on LSTM -- with two hidden layers -- called DeepTrend is presented in \cite{Dai2017}. The extraction layer, a fully connected layer, learns the time variant trend and the prediction layer, a LSTM layer, performs the forecast based on the extracted trend and the calculated residual series. The introduced model was compared to other ML and classical time-series approaches. DeepTrend achieved the lowest error value on the PeMS dataset which was restricted to 16 weeks of 2016 and 50 stations in district 4. Regrettably, the training and testing time is unobserved during the comparison.

In \cite{Zhang2018} a model is introduced based on gated recurrent units (GRU) which incorporates weather and traffic data. The model was tested on different datasets and compared with a variety of ML models. Considering the loss and accuracy, the proposed model achieved the best values. The prediction accuracy increased with the data fusion of weather conditions and traffic flow. Unfortunately, the training and testing duration is omitted. 

Moreover, in \cite{Huang2014} a deep belief network (DBN) is proposed. The DBN configuration is done by random search by constraining the searchspace. The DBN is applied on two datasets, including the PeMS dataset. Different scenarios are considered, e.\,g. grouped input data as well as a variety of forecast horizons. The DBN showed in all experiences superiors results regarding the error value. 

In \cite{Polson2017} a DL model is proposed, its hyperparameter and structure became tuned by random search. The search space was heavily constrained and the model configurations were selected by a monte carlo algorithm. Different data pre-processing approaches were considered and utilized for comparing a linear, shallow and DL model. The best results were achieved by the DL model with median filter pre-processing and L1 regularization. Regrettably, the model complexity as well as the training and inference duration are not examined. 

Additionally, in \cite{Fu2017} autoregressive integrated moving average (ARIMA) as well as  LSTM and GRU neural networks (NNe are evaluated. The models were applied to a part of the PeMS dataset. The prediction results for LSTM and GRU NN were reported as similar and both better than ARIMA. The error values achieved by the GRU NN were lower than those stated for the LSTM NN. The architectural details of the recurrent NNs (RNNs) were omitted. Moreover, the training and inference duration were neglected.

In \cite{Wei2019} a model based on AutoEncoder and LSTM NNs is proposed. The investigations were performed on the PeMS dataset. Each network is preliminary separately trained whereby the learned representation of the AutoEncoder NN is concatenated with the measured traffic data and used for the training of the LSTM NN. Finally, both NNs are trained together for a fine adjustment of the weights. The introduced model achieved the lowest error rates compared to several ML approaches and various forecast horizons. Unfortunately, the training and inference duration are left unconsidered.

Besides, in \cite{Chen2018} a model based on a generative adversarial network (GAN) and stacked LSTM is introduced. The PeMS dataset  -- district 5 -- of 2013 was examined and synthetically extended by the developed GAN. Both, the real and generated traffic flow data were used for training the LSTM. A comparison of the LSTM trained on both, the synthetic and real as well as only on real data with different historical time steps leads to the conclusion, that the smallest error values are achieved by the proposed model. The training and inference duration are omitted.

In this paper we report about various RNNs utilized to forecast the traffic flow in the city of Hagen. After training the proposed models with traffic data over 6 months, the RNNs can be used to predict the traffic flow at each sensor location simultaneously for the configured forecast horizon. 

Thus, the trained RNNs can be considered as structure-independent models of Hagen’s street network, so that we neither need hand-crafted street graphs that map the underlying streets nor have to make complicated assumptions about the drivers behavior. Instead, we provide the current and historical traffic data of sensors from crossings to the RNNs to perform the traffic forecasts. 

Our contribution is manifold: 1) Handling missing data and pre-processing; 2) Comparison of various models; 3) Relation between GRU and LSTM cells.

The paper is organized as follows: Section 2 is dedicated to the data description and pre-processing. In Section 3 we introduce the utilized RNNs as well as the implementation details. In Section 4 the results are presented and discussed thoroughly. Finally, we close with a conclusion and an outlook on future works.

%% file: section2/section2.tex
\section{Data analysis and preprocessing}
All measurements are sent to and stored in a central traffic control computer which is also able to control the traffic lights. The recorded data contains measurements from 129 inductive sensors at 12 intersections. Other works often focused on freeway data \cite{Dai2017,Lippi2013}. 
The provided measurements were stored every minute in the unit \textit{vehicles per hour $\frac{veh}{h}$}, which indicates how many vehicles in average were passing that sensor in the proceeding hour. Most often, the traffic control computer uses steps of 60 $\frac{veh}{h}$ to get an integer result when the value is divided by 60 minutes. This results in inaccurate measurements and a high fluctuation over time in the data. The measured values from 6\,am to 10\,pm of every weekday were analysed and used for forecasting.
It turned out that some of the sensors never stored data respectively no other values than 0 $\frac{veh}{h}$ over a longer period. These sensors were unusable for forecasting traffic flow. There were also sensors which did not provide measurements every minute so that the number of available values per minute fluctuates very strongly between 56 and 129 during the period of record.

At last 108 usable series of measurements from sensors that provided data over all or respectively most of the given time range. Because there were still many missing values in the time series, this work takes into account how missing data should be handled, before they were processed by RNNs. The following three approaches were considered: 1) Missing values were marked with $-1$. This set will be considered as raw data. 2) Missing values were repaired by inserting averaged values from all identical weekdays at the same record time. Linear interpolation was used if this was not possible, in the following considered as repaired data. 3) The data was aggregated per driving direction (usually 3 to 4 directions per intersection), so that an average value was calculated for all lanes into the same direction, based on the repaired data. This set will be considered as aggregated data. The datasets were divided so that 87\% was used for training, 3\% for validation and 10\% for test. This segmentation assured that the training data as well as the test data contains also weekends and holidays, because of the different amount of traffic on the streets during these days.

\begin{figure}[bt]
	\centering
    \begin{minipage}[t]{0.45\linewidth}
        \centering
        \includegraphics[width=1\linewidth]{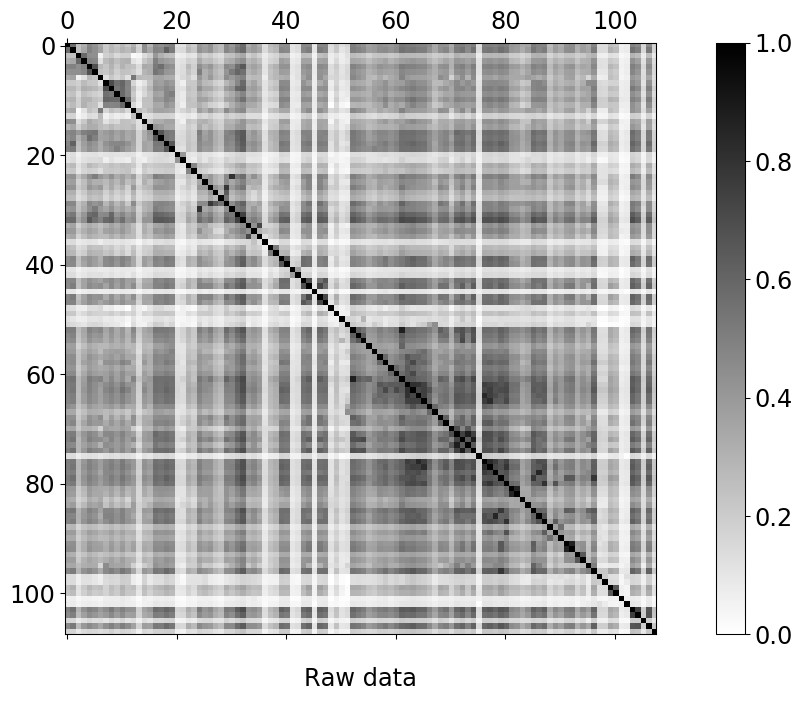}
    \end{minipage}
    \hfill
    \begin{minipage}[t]{0.45\linewidth}
        \centering
        \includegraphics[width=\linewidth]{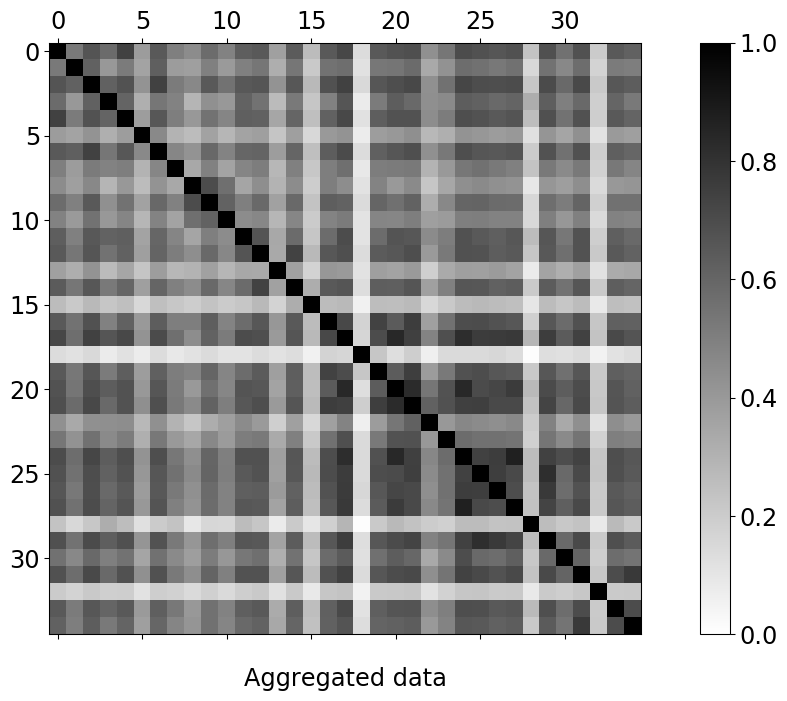}
    \end{minipage}
	\caption{Correlation matrices of sensor time series. Sensors and driving directions are identified by numbers from 0 to 107 respectively 0 to 34.}
	\label{fig:Korrelation}
\end{figure}

Due to, the localisation of the sensors inside the city centre, we assumed that multiple intersection were passed by the same vehicles resulting in relations between the measurements of different sensors. Sometimes, multiple sensors are on the same lane at an intersection, which differs from the PeMS dataset where in general only one is present \cite{Lippi2013}. Obliviously, it could be beneficial to process all data at the same time so that the relation of the data is used for the forecast. This assumption was proofed by calculating correlation matrices illustrating the correlation between different time series of the sensors. The result is presented in Fig. \ref{fig:Korrelation} and shows that there is a distinct correlation between many sensors. The matrix of the repaired data is omitted because it had shown similar results as the raw data.
This work had also analysed multiple step sizes for forecasting. E.g. when a step size of 15 min. was used, the given data was resampled over 15 min. and considered as one averaged value before it was used to train the NNs.

%% file: section3/section3.tex
\section{RNN Architectures and Implementation Details}

This work analysed the accuracy of forecasting traffic flow with RNNs by using different types and architectures. All resulting models were implemented by using \texttt{TensorFlow 1.12} \cite{tensorflow2015-whitepaper} with \texttt{CUDA} support and the \texttt{Scientific Computing Tools for Python (SciPy)} on \texttt{Python 3}. The evaluation of the models was conducted on a mid-level gaming graphic card -- Nvidia GTX 1060 -- and on professional computing GPUs -- Nvidia P100 and V100 -- in Google Cloud Virtual Machines. 

\subsection{Models}
Based on well-proven models for time series prediction, three architectures were selected for traffic flow prediction \cite{Cirstea2018,machinelearningmastery,Harmon2018}. Each Architecture was instantiated as a model with LSTM and GRU layers as recurrent component, which resulted in six models. Every model was designed in the way that it took sequences of measurements of each sensor as input and generated sequences of predictions as output for every sensor at the same time. 
Input and output length had not to have necessarily the same length, so that different combinations of in- and output length could be analysed. 
In the following, a recurrent layer can be either a LSTM or a GRU layer. Refer to the relevant documentation for further details especially how the data is processed internally \cite{tensorflow2015-whitepaper}.

A \textbf{Convolutional Recurrent Neural Network (CRNN)} is a combination of a convolutional neural network and a RNN. The used CRNN architecture consists of two convolutional layers to extract relevant features from the measured traffic data by convolving the input over a single spatial dimension. The most relevant features are extracted by a subsequent maximum pooling layer. The data, which is still a sequence at this point, is now flattened to a large vector by a flatten layer. The vector will be repeated by a repeat layer according to the used prediction length. The repeat layer is followed by two recurrent layers which generate sequences of predicted data based on the repeated input. Because the output of the last recurrent layer is still an internal representation, there are two following dense layers which convert the internal representation into sequences for each sensor. To avoid overfitting, a dropout layer after the pooling layer and spatial dropout layer after each recurrent layer were used.

The \textbf{Encoder-Decoder} architecture uses an encoder to generate an internal representation of the data. The decoder forms the internal representation into the original format respectively into a new desired format. The latter is used to have a different length for input and output sequences. There are also hidden layer for the actual forecasting next to the encoder and decoder. The developed architecture used a recurrent layer to transform the input sequences from each sensor into an internal static representation of the input. This internal state is repeated according to the prediction length by a repeat layer and feed into a second recurrent layer for forecasting. The internal prediction is than decoded by two dense layer into the output format so that it consists of predicted time series for every sensor. A dropout layer was inserted after the first recurrent layer to reduce overfitting. Because the first recurrent layer is configured to not produce sequences, a non-sequential dropout layer is used.

The \textbf{Vector-Output} architecture generates a vector with predicted time series for every sensor at its output without repeating any internal states. This architecture uses two recurrent layers at its input to produce predicted time series directly. The first recurrent layer takes sequences as input and also produces sequences for each sensor at its output which are forwarded to the second recurrent layer. The second layer computes a static output of all input sequences. This is further processed by a dense layer into a large vector containing concatenated time series with the predicted traffic flow for each sensor. This vector is transformed into the correct dimension by a reshape layer. This forms the vector into time series for each sensor. A dropout layer was inserted after the second recurrent layer which also does not produce sequences.

\subsection{Training and evaluation}
\label{sec:evaluation}
The recorded time series were fed into the models together with the corresponding weekdays (0 to 6) and timestamps (0 to 959). Each time series might be resampled according to the current step size before all values were scaled by using a \texttt{RobustScaler} \cite{scikit-learn} to avoid exploding and vanishing gradients \cite{Bengioa}. The training data was used in a sequential order to extract sequences of measurements to train all models  which resulted in more accurate forecasts, due to the sequential order of the data given by nature \cite{LeCun2012}. All models were fine tuned regarding the available hyperparameter to fit a specific combination of input length, step size, prediction length and data set by applying grid search. Every model utilized the mean squared error loss function together with the adam optimizer \cite{Adam2017}. Furthermore, L2 regularisation was used to reduce overfitting.

To evaluate the best model, best length of forecasting traffic flow and the necessary pre-computation of the data, every model was trained and tested with all three data sets, step sizes of 1, 5, 15, 30 and 60 min. as well as prediction lengths of 1, 5 and 20 steps. This resulted in 270 training and evaluation phases. The training was always computed for a fixed amount of 300 epochs, each comprising 200 steps respectively input sequences of all sensors. All models were fine tuned for the following combination: Input length: 50, step size: 5 min., prediction length: 5, data set: repaired data.

%% file: section4/section4.tex
\section{Results and Discussion}

This section outlines the results of forecasting the traffic flow with the proposed RNNs. It starts with an overview of the results achieved by the developed architectures and is finalized by implications from the results.

\subsection{Model evaluation}
All models were trained and tested with the in Sec. \ref{sec:evaluation} listed combinations. During the test phase the Root Mean Square Error (RMSE) and the coefficient of determination ($R^2$) over all prediction steps were calculated \cite{scikit-learn}. The results of every model were compared for each combination of step size and prediction length and are shown in \cref{tab:Results_P1,tab:Results_P5,tab:Results_P20}. 

\input{section4/resulttables}

Every table contains the results for one of the three tested prediction lengths which are separated by datasets. For every architecture as well as for each step size the RMSE and $R^2$ are stated. The best RMSE for each combination is highlighted by bold printing.
The results in Table \ref{tab:Results_P1} show a big difference in model performance for the step sizes of 1 and 5 min. A detailed analysis was done with a CRNN-LSTM model which showed especially for a step size of 1 min., that the measured traffic has a high fluctuation where the models predicted the average traffic flow, caused by a too low complexity of the used models. Obviously, the high fluctuation decreased significantly by averaging the measured traffic flow over 5 min. With this input data the model already performed significantly better, even if steep peaks are still not satisfactory predicted.
The results in the table also shows that the forecasting error for the raw data becomes lowest starting with a step size of 15 min. It should be noted that in case of a sensor defect, this dataset will be unusable to do forecasting for this sensor. The aggregated data instead would be still able to do meaningful forecast for a specific direction. Furthermore, RMSE and $R^2$ are changing as expected where the RMSE decreases and $R^2$ increases with increasing step sizes. 
Table \ref{tab:Results_P5} shows the results for a prediction length of five steps. The results are competitive comparable to the results in Tab. \ref{tab:Results_P1} except of some outliers.
All results for a prediction length of 20 steps are listed in Tab. \ref{tab:Results_P20}. The results show no significant difference compared to the results in \cref{tab:Results_P1,tab:Results_P5}, especially for larger step sizes where the data is averaged over a longer period of time. Moreover, the results for the CRNN architecture are closer to them scored by the other architectures than for the shorter prediction lengths. It should be noted that there was not enough data for a step size of 60 min. and a prediction length of 20 steps to use a input length of 200 values than before for all other step sizes. It was shortened to 100 values to be able to test a step size of 60 min.

The Vector-Output model with GRU performed best regarding RMSE and $R^2$. The results show that the RMSE for the different prediction length did not increase more than 5 $\frac{vec}{h}$ between 1 and 20 forecasting steps and step sizes larger than one min. Besides, the results state that averaging high fluctuating sensor data improves the forecasting quality dramatically. The comparison of the results scored with the different datasets presented that the RMSE with the repaired data were often worse than the one calculated with raw data and that the aggregated data leads to similar results as the raw data.

\begin{figure}[!b]
	\centerline{\includegraphics[width=1\textwidth]{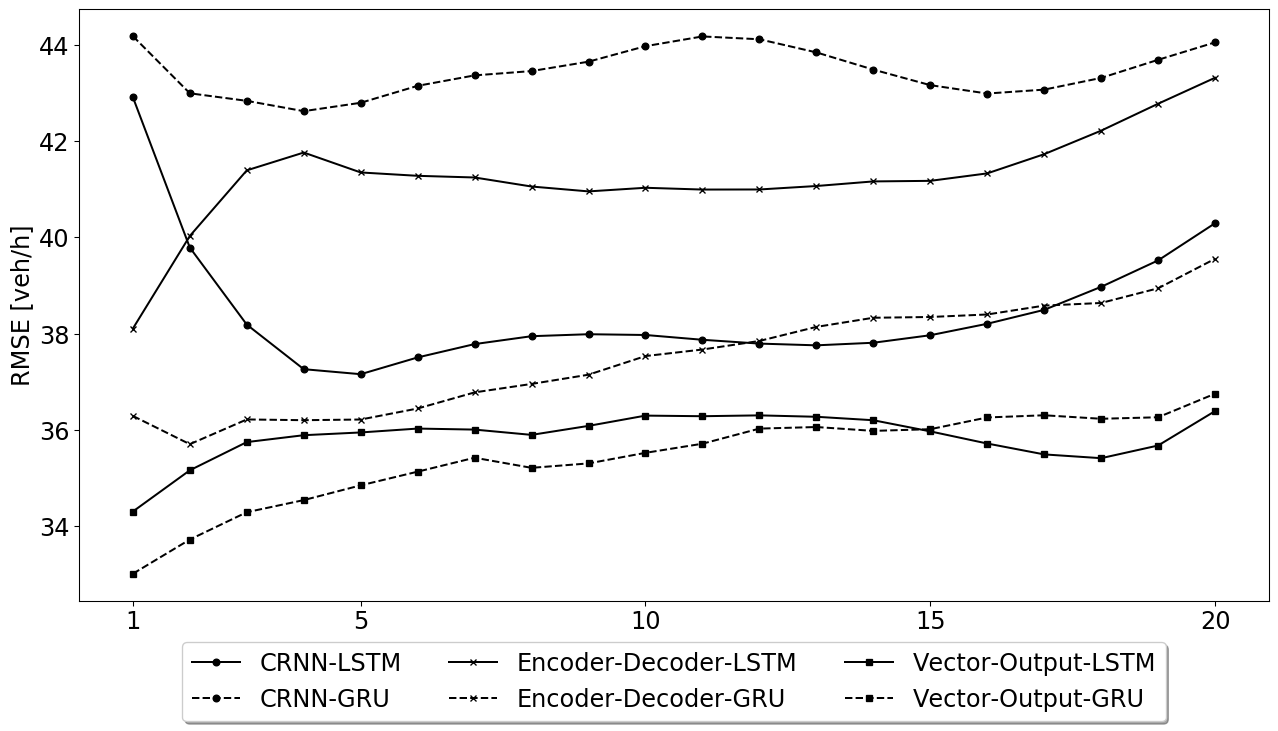}}
	\caption{Comparison of the models at forecasting 20 steps with a step size of 30 min. while using raw data.}
	\label{fig:PredictionLength20}
\end{figure}

Derived from the given results, tagging missing measurements from sensors is sufficient for the traffic flow prediction. Additionally, forecasts of multiple steps did achieve considerably similar error values. The given sensor data requires to be averaged to reduce the before mentioned drawbacks of the steps of 60 $\frac{vec}{h}$. Hence, it is beneficial to use a step size of 30 min., the necessary time between two adjustments of traffic lights in the city of Hagen. Figure \ref{fig:PredictionLength20} shows the model results of forecasting 20 steps with a step size of 30 min. by utilizing raw data. 

Especially both Vector-Output models show only a small deviation between the lowest and highest error. The RMSE is sometimes decreasing with further prediction steps. All models were trained and tested with an input of sequences containing 200 aggregated measurements for each sensor. The number of weights for each model as well as training and testing duration were listed in Tab. \ref{tab:Training_P20_W30}. 

\begin{table}[ht]\caption{Number of weights and timings for training and forecasting for each model to predict 20 steps with a step size of 30 min. by using raw data.} 

\centering
\begin{tabular}{|p{4cm}|>{\raggedleft\arraybackslash}p{2.5cm}|>{\raggedleft\arraybackslash}p{2.5cm}|>{\raggedleft\arraybackslash}p{2.5cm}|}
\hline
	\textbf{Model}  & \multicolumn{1}{|c|}{\textbf{Time to train}}                    & \multicolumn{1}{|c|}{\textbf{Time to}}                 &\multicolumn{1}{|c|}{\textbf{Number of}} \\
	                & \multicolumn{1}{|c|}{\textbf{[mm:ss]}}    & \multicolumn{1}{|c|}{\textbf{forecast [ms]}}    & \multicolumn{1}{|c|}{\textbf{weights}} \\
\hline
    CRNN-LSTM				& 12:52 & 4.2 & 5,135,616 \\
	CRNN-GRU				& 11:34	& 3.9 & 3,907,440 \\
	Encoder-Decoder-LSTM	& 19:53	& 3.7 & 447,012 \\
	Encoder-Decoder-GRU		& 18:51	& 3.6 & 352,836 \\
	Vektor-Ausgabe-LSTM		& 53:10	& 3.8 & 2,341,872 \\
	Vektor-Ausgabe-GRU		& 49:03	& 3.7 & 1,990,224 \\
\hline
\end{tabular}
\label{tab:Training_P20_W30} 
\end{table}

Timings were measured on the professional compute card. It shows that the CRNN models have the highest number of weights and was trained as fastest. The Encoder-Decoder models required around twice the time for the training although they have a smaller amount of weights. Compared to the CRNN, the training time for the Vector-Output model, with about half of the number of weights, increased four times. Obviously, the number of weights itself is not essential for the compute time but their distribution. This is because the CRNN models had around 95\% of their weights inside the recurrent layers where \texttt{CUDA} optimized versions were used. Rather the Vector-Output models only had 60\% respectively 53\% of their weights inside recurrent layers.

Due to the lower complexity of the GRU cells, the GRU versions of each architecture was always faster than the LSTM version. The inference duration of each model was nearly the same ($\approx$ 4\,ms). The GRU variants were always a little faster than the corresponding LSTM once.

\subsection{Implications}
We showed that good results in forecasting traffic flow were achieved with multivariate regression for 108 sensors. The best model showed a total RMSE of 31 $\frac{vec}{h}$ in presence of peak loads of more than 2000 $\frac{vec}{h}$ at some of the sensors.
Further analysis of the local RMSE at each sensor showed that some of them had a relatively high error compared to others. This means that the traffic flow at those sensors is harder to predict than for others. This could be related to irregular loads on these streets, but further analysis is needed.

It should also be noted that every model was optimized at only one input combination (refer to section \ref{sec:evaluation}) and was not fine tuned for the combination determined as best choice. Further fine tuning of the models hyperparameter could result in lower forecasting errors. Furthermore, the statistic stability of training NNs with randomly initialized weights was not considered during the evaluation of the models. This means, that every model was trained and tested only once due to the long computing times. An analysis of training the same model 20 times was made with the Vector-Output model with the combination mentioned as best. The results presented that 50\% of every prediction step of every run had a RMSE deviation smaller than 3 $\frac{vec}{h}$ and that the median RMSE of all forecasts between step 1 and 20 showed an increase of 5.5 $\frac{vec}{h}$.

%% file: section4/resulttables.tex
\begin{table}\caption{\small Results for predicting one step. RMSE is  given in $\frac{vec}{h}$.} 

\scriptsize
\centering
\begin{adjustbox}{max width=\textwidth}
\begin{tabular}{|l|ll|rr|rr|rr|rr|rr|}
\hline
  \multirow{2}{*}{\textbf{Data set}} & \multicolumn{2}{c|}{\textbf{Step size}} & \multicolumn{2}{c|}{\textbf{1 minute}} & \multicolumn{2}{c|}{\textbf{5 minutes}} & \multicolumn{2}{c|}{\textbf{15 minutes}} & \multicolumn{2}{c|}{\textbf{30 minutes}} & \multicolumn{2}{c|}{\textbf{60 minutes}} \\
\cline{2-3}
	 &  \multicolumn{2}{c|}{\textbf{Model}} &      RMSE &      $R^2$ &     RMSE &      $R^2$ &     RMSE &     $R^2$ &     RMSE &     $R^2$ &     RMSE &     $R^2$ \\
\hline
	\multirow{6}{*}{Raw} 
    & \multirow{2}{*}{CRNN} & LSTM &  122.6 &  0.18 &    58.3 &  0.53 &     41.0 &   0.70 &     36.3 &   0.76 &     30.5 &   0.81 \\
    &              	& GRU &  123.1 &  0.18 &    56.8 &  0.55 &     41.4 &   0.70 &     35.5 &   0.77 &     30.4 &   0.82 \\
\cline{2-13}
    & \multirow{2}{*}{Encoder-Decoder} & LSTM &   96.9 &  0.39 &    53.7 &  0.59 &     39.7 &   0.73 &     32.8 &   0.80 &     28.9 &   0.84 \\
    &    					 	& GRU &   96.5 &  0.39 &    \textbf{52.8} &  0.60 &     39.1 &   0.74 &     33.4 &   0.80 &     30.4 &   0.83 \\
\cline{2-13}
    & \multirow{2}{*}{Vector-Output} & LSTM &   \textbf{96.3} &  0.38 &    53.3 &  0.60 &     38.8 &   0.74 &     33.9 &   0.78 &     28.8 &   0.84 \\
    &    						& GRU &  99.0 &  0.38 &    53.0 &  0.60 &     \textbf{38.3} &   0.74 &     \textbf{31.4} &   0.81 &     \textbf{28.0} &   0.84 \\
 \hline
 \multirow{6}{*}{Repaired} 
  & \multirow{2}{*}{CRNN} & LSTM &  114.3 &  0.18 &    69.3 &  0.53 &     55.0 &   0.70 &     49.2 &   0.74 &     42.2 &   0.81 \\
  &             		& GRU &  113.7 &  0.18 &    63.8 &  0.55 &     54.4 &   0.70 &     47.9 &   0.76 &     41.2 &   0.82 \\
\cline{2-13}
  & \multirow{2}{*}{Encoder-Decoder} & LSTM &  103.4 &  0.30 &    60.5 &  0.60 &     \textbf{49.1} &   0.74 &     42.4 &   0.79 &     37.9 &   0.84 \\
  &   						 	& GRU &  102.2 &  0.30 &    \textbf{59.6} &  0.61 &     49.2 &   0.73 &     44.7 &   0.77 &     \textbf{37.2} &   0.84 \\
\cline{2-13}
  & \multirow{2}{*}{Vector-Output} & LSTM &  \textbf{101.8} &  0.31 &    60.2 &  0.60 &     50.4 &   0.74 &     44.0 &   0.79 &     38.4 &   0.83 \\
  &     						& GRU &  105.3 &  0.28 &    60.0 &  0.60 &     49.6 &   0.73 &     \textbf{42.2} &   0.79 &     38.4 &   0.83 \\
 \hline 
 \multirow{6}{*}{Aggregated}
   & \multirow{2}{*}{CRNN} & LSTM &   70.0 &  0.52 &    52.6 &  0.67 &     43.3 &   0.76 &     37.7 &   0.81 &     34.4 &   0.82 \\
   &             		& GRU &  69.7 &  0.52 &    51.4 &  0.68 &     42.7 &   0.77 &     37.1 &   0.81 &     33.1 &   0.83 \\
\cline{2-13}
   & \multirow{2}{*}{Encoder-Decoder} & LSTM &   56.8 &  0.63 &    45.9 &  0.73 &     39.1 &   0.81 &     34.4 &   0.85 &     29.6 &   0.88 \\
   &   							& GRU &  55.5 &  0.64 &    \textbf{45.0} &  0.73 &     39.8 &   0.81 &     34.8 &   0.84 &     31.8 &   0.86 \\
\cline{2-13}
   & \multirow{2}{*}{Vector-Output} & LSTM &   \textbf{54.5} &  0.66 &    45.4 &  0.73 &     40.0 &   0.81 &     34.1 &   0.84 &     29.8 &   0.88 \\
   &     						& GRU &   55.4 &  0.65 &    46.0 &  0.72 &     \textbf{39.1} &   0.81 &     \textbf{33.3} &   0.85 &     \textbf{27.5} &   0.90 \\
\hline
\end{tabular}
\end{adjustbox}
\label{tab:Results_P1} 

\caption{\small Results for predicting five steps. RMSE is  given in $\frac{vec}{h}$.} 
\begin{adjustbox}{max width=\textwidth}
\begin{tabular}{|l|ll|rr|rr|rr|rr|rr|}
\hline
  \multirow{2}{*}{\textbf{Data set}} & \multicolumn{2}{c|}{\textbf{Step size}} & \multicolumn{2}{c|}{\textbf{1 minute}} & \multicolumn{2}{c|}{\textbf{5 minutes}} & \multicolumn{2}{c|}{\textbf{15 minutes}} & \multicolumn{2}{c|}{\textbf{30 minutes}} & \multicolumn{2}{c|}{\textbf{60 minutes}} \\
\cline{2-3}
	 &  \multicolumn{2}{c|}{\textbf{Model}} &      RMSE &      $R^2$ &     RMSE &      $R^2$ &     RMSE &     $R^2$ &     RMSE &     $R^2$ &     RMSE &     $R^2$ \\
\hline
	\multirow{6}{*}{Raw} 
    & \multirow{2}{*}{CRNN} & LSTM &  124.8 &  0.17 &    59.2 &  0.52 &     44.6 &   0.67 &     35.9 &   0.77 &     31.2 &   0.81 \\
    &              	& GRU &  125.6 &  0.16 &    59.8 &  0.52 &     43.2 &   0.68 &     36.9 &   0.76 &     32.9 &   0.80 \\
\cline{2-13}
    & \multirow{2}{*}{Encoder-Decoder} & LSTM &   123.6 &  0.18 &    55.6 &  0.58 &     41.4 &   0.70 &     35.9 &   0.77 &     39.5 &   0.75 \\
    &    					 	& GRU &   97.0 &  0.38 &    55.2 &  0.58 &     40.5 &   0.69 &     35.9 &   0.78 &     32.6 &   0.82 \\
\cline{2-13}
    & \multirow{2}{*}{Vector-Output} & LSTM &  \textbf{95.8} &  0.39 &    55.1 &  0.58 &     41.6 &   0.71 &     37.0 &   0.75 &     30.6 &   0.83 \\
    &    						& GRU &  121.2 &  0.20 &    \textbf{54.4} &  0.59 &     \textbf{40.5} &   0.73 &     \textbf{34.1} &   0.78 &     \textbf{27.8} &   0.86 \\
 \hline
 \multirow{6}{*}{Repaired} 
  & \multirow{2}{*}{CRNN} & LSTM &  113.5 &  0.19 &    68.6 &  0.56 &     56.9 &   0.69 &     46.4 &   0.78 &     46.7 &   0.77 \\
  &             		& GRU &  144.9 & -0.29 &    68.4 &  0.57 &     55.5 &   0.70 &     \textbf{47.5} &   0.76 &     45.6 &   0.78 \\
\cline{2-13}
  & \multirow{2}{*}{Encoder-Decoder} & LSTM & 105.1 &  0.28 &    66.4 &  0.57 &     53.0 &   0.72 &     53.3 &   0.69 &     45.2 &   0.76 \\
  &   						 	& GRU &  107.9 &  0.24 &    67.1 &  0.58 &     \textbf{51.6} &   0.73 &     58.3 &   0.63 &     \textbf{40.8} &   0.82 \\
\cline{2-13}
  & \multirow{2}{*}{Vector-Output} & LSTM &  \textbf{102.2} &  0.31 &    \textbf{65.8} &  0.59 &     54.4 &   0.67 &     48.4 &   0.77 &     42.3 &   0.82 \\
  &     						& GRU &  102.2 &  0.30 &    66.6 &  0.58 &     53.1 &   0.73 &     47.7 &   0.77 &     42.4 &   0.80 \\
 \hline 
 \multirow{6}{*}{Aggregated}
   & \multirow{2}{*}{CRNN} & LSTM &   70.8 &  0.51 &    54.5 &  0.66 &     46.4 &   0.73 &     42.3 &   0.76 &     44.3 &   0.73 \\
   &             		& GRU &   70.5 &  0.51 &    53.0 &  0.67 &     46.3 &   0.72 &     39.7 &   0.79 &     39.5 &   0.79 \\
\cline{2-13}
   & \multirow{2}{*}{Encoder-Decoder} & LSTM &   64.1 &  0.57 &    51.7 &  0.69 &     45.1 &   0.76 &     40.4 &   0.80 &     33.9 &   0.86 \\
   &   							& GRU &   \textbf{63.6} &  0.57 &    50.8 &  0.69 &     44.0 &   0.76 &     43.0 &   0.77 &     33.5 &   0.85 \\
\cline{2-13}
   & \multirow{2}{*}{Vector-Output} & LSTM &  63.9 &  0.57 &    \textbf{50.3} &  0.70 &     \textbf{40.2} &   0.80 &     36.7 &   0.83 &     37.8 &   0.82 \\
   &     						& GRU &  63.7 &  0.57 &    51.6 &  0.69 &     44.3 &   0.77 &     \textbf{36.5} &   0.83 &     \textbf{33.3} &   0.85 \\
\hline
\end{tabular}
\end{adjustbox}
\label{tab:Results_P5} 

\caption{\small Results for predicting twenty steps. RMSE is  given in $\frac{vec}{h}$.} 
\begin{adjustbox}{max width=\textwidth}
\begin{tabular}{|l|ll|rr|rr|rr|rr|rr|}
\hline
  \multirow{2}{*}{\textbf{Data set}} & \multicolumn{2}{c|}{\textbf{Step size}} & \multicolumn{2}{c|}{\textbf{1 minute}} & \multicolumn{2}{c|}{\textbf{5 minutes}} & \multicolumn{2}{c|}{\textbf{15 minutes}} & \multicolumn{2}{c|}{\textbf{30 minutes}} & \multicolumn{2}{c|}{\textbf{60 minutes}} \\
\cline{2-3}
	 &  \multicolumn{2}{c|}{\textbf{Model}} &      RMSE &      $R^2$ &     RMSE &      $R^2$ &     RMSE &     $R^2$ &     RMSE &     $R^2$ &     RMSE &     $R^2$ \\
\hline
	\multirow{6}{*}{Raw} 
    & \multirow{2}{*}{CRNN} & LSTM &  118.2 &  0.26 &    64.5 &  0.47 &     44.2 &   0.69 &     38.5 &   0.74 &     41.8 &   0.69 \\
    &              	& GRU & 120.3 &  0.24 &    61.8 &  0.50 &     45.8 &   0.67 &     43.5 &   0.68 &     34.2 &   0.79 \\
\cline{2-13}
    & \multirow{2}{*}{Encoder-Decoder} & LSTM &   121.1 &  0.17 &    57.6 &  0.55 &     44.0 &   0.70 &     41.3 &   0.72 &     48.2 &   0.60 \\
    &    					 	& GRU &   122.7 &  0.18 &    57.8 &  0.55 &     45.1 &   0.67 &     37.5 &   0.75 &     38.8 &   0.73 \\
\cline{2-13}
    & \multirow{2}{*}{Vector-Output} & LSTM &  \textbf{116.5} &  0.23 &    55.9 &  0.58 &     44.8 &   0.69 &     35.9 &   0.77 &     38.4 &   0.75 \\
    &    						& GRU &  127.7 &  0.17 &    \textbf{55.6} &  0.58 &     \textbf{42.0} &   0.71 &     \textbf{35.4} &   0.78 &     \textbf{30.3} &   0.83 \\
 \hline
 \multirow{6}{*}{Repaired} 
  & \multirow{2}{*}{CRNN} & LSTM &  106.8 &  0.27 &    71.7 &  0.54 &     61.6 &   0.63 &     54.0 &   0.72 &     51.5 &   0.74 \\
  &             		& GRU & 107.6 &  0.26 &    74.9 &  0.49 &     65.1 &   0.61 &     56.1 &   0.69 &     49.4 &   0.77 \\
\cline{2-13}
  & \multirow{2}{*}{Encoder-Decoder} & LSTM &  \textbf{101.4} &  0.31 &    74.1 &  0.52 &     55.7 &   0.71 &     53.5 &   0.72 &     49.3 &   0.76 \\
  &   						 	& GRU &  102.6 &  0.31 &    68.5 &  0.56 &     56.0 &   0.69 &     53.4 &   0.70 &     48.9 &   0.76 \\
\cline{2-13}
  & \multirow{2}{*}{Vector-Output} & LSTM &  108.5 &  0.26 &    \textbf{68.0} &  0.57 &     61.0 &   0.68 &     51.0 &   0.76 &     48.7 &   0.78 \\
  &     						& GRU & 123.5 & -0.17 &    68.3 &  0.57 &     \textbf{53.7} &   0.72 &     \textbf{50.6} &   0.75 &     \textbf{46.1} &   0.81 \\
 \hline 
 \multirow{6}{*}{Aggregated}
   & \multirow{2}{*}{CRNN} & LSTM &   72.3 &  0.50 &    57.3 &  0.62 &     53.6 &   0.66 &     50.1 &   0.67 &     41.6 &   0.75 \\
   &             		& GRU &   72.6 &  0.50 &    58.2 &  0.61 &     60.0 &   0.57 &     45.4 &   0.73 &     41.4 &   0.76 \\
\cline{2-13}
   & \multirow{2}{*}{Encoder-Decoder} & LSTM &  70.2 &  0.53 &    54.5 &  0.66 &     53.1 &   0.67 &     42.1 &   0.76 &     32.9 &   0.85 \\
   &   							& GRU &  69.0 &  0.53 &    \textbf{54.5} &  0.66 &     45.1 &   0.76 &     40.5 &   0.79 &     34.0 &   0.85 \\
\cline{2-13}
   & \multirow{2}{*}{Vector-Output} & LSTM &  70.0 &  0.53 &    56.2 &  0.64 &     \textbf{42.8} &   0.78 &     \textbf{36.0} &   0.83 &     34.4 &   0.83 \\
   &     						& GRU &  \textbf{68.9} &  0.53 &    54.6 &  0.66 &     44.1 &   0.76 &     37.9 &   0.81 &     \textbf{32.7} &   0.85 \\
\hline
\end{tabular}
\end{adjustbox}
\label{tab:Results_P20} 
\end{table}

%% file: section5/section5.tex
\section{Conclusion and Future Works}

In this paper we analyzed forecasting the traffic flow in the city of Hagen with RNNs. The data was generated by inductive loop sensors distributed around the city centre. Moreover, the data became pre-processed, to incorporate spatial dependencies and handle missing data. We developed three RNN architectures and evaluated them as instances with LSTM and GRU cells on different input data and forecasting horizons. We showed that forecasting the traffic flow inside cities can be done satisfactory by RNNs, even for large forecast horizons. Our best model, the Vector-Output architecture with GRU cells, predicted the traffic 10 hours into the future by considering measurements every 30 minutes. The forecasts at 108 sensors have been performed simultaneously with a total error of 35 $\frac{vec}{h}$ at peak loads of more than 2000 $\frac{vec}{h}$ at some sensors. Besides, we showed that computing all sensors at the same time is beneficial because the measurements are correlated between sensors inside the city. We discovered that it is sufficient to use the sensors raw data and to tag missing measurements inside the data set.

Due to limited data, the model will be trained and tested with more data to increase generalization effects. Furthermore, the hyperparameters will be fine tuned to fit the determined combination of step size, prediction length and data set. Additionally, we will take statistical deviations regarding weight initialization during model evaluation into account. It emerged that some of the sensors are harder to predict than others during the evaluation phase. Further studies are necessary to identify the root cause and to determine how these effects could be mitigated. In addition to that we will implement the developed model in our smart mobility app called STREAM (Smart TRaffic using Edge And social CoMputing) with the objective of a better, more balanced utilization of the street network.